%% file: main.tex
\documentclass{article}

\PassOptionsToPackage{numbers, square}{natbib}

\usepackage[final]{neurips_2025}

\usepackage[utf8]{inputenc} 
\usepackage[T1]{fontenc}    
\usepackage{hyperref}       
\usepackage{url}            
\usepackage{booktabs}       
\usepackage{amsfonts}       
\usepackage{amsmath}        
\usepackage{nicefrac}       
\usepackage{microtype}      
\usepackage{natbib}
\usepackage{graphicx}
\usepackage{multirow}
\usepackage{multicol}
\usepackage{enumitem}
\usepackage{wrapfig}
\usepackage{bm}
\usepackage{pifont}
\usepackage[table]{xcolor}

\title{
\raisebox{-0.25\height}{\includegraphics[width=0.7cm]{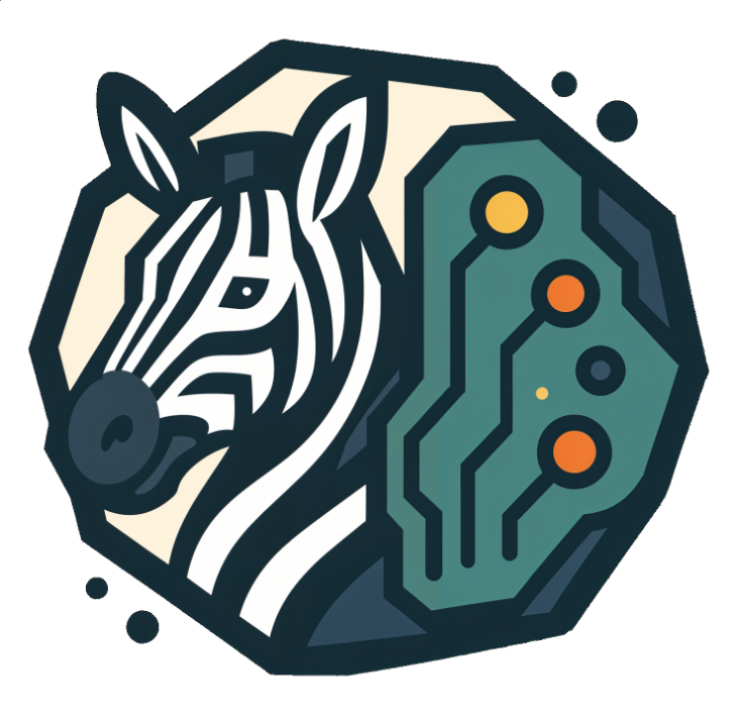}}
\methodname{}: Towards \underline{Ze}ro-Shot Cross-Subject Generalization for Universal \underline{Bra}in Visual Decoding
}

\author{%
  Haonan Wang,
  Jingyu Lu, 
  Hongrui Li,
  Xiaomeng Li\thanks{Corresponding author.} \\
  The Hong Kong University of Science and Technology\\
  \texttt{hwanggr@connect.ust.hk, eexmli@ust.hk} \\
}

\newcommand{\methodname}{\textsc{Zebra}}

\begin{document}

\maketitle

\input{sec/0_abstract}    
\input{sec/1_intro}

\input{sec/2_related_works}

\input{sec/3_method}
\input{sec/4_exp}

\input{sec/5_conclusion}

\setlength{\bibsep}{0em}
\bibliographystyle{ieeetr}
{
\small
\bibliography{main}
}

\input{sec/checklist}

\end{document}

%% file: sec/0_abstract.tex
\begin{abstract}
Recent advances in neural decoding have enabled the reconstruction of visual experiences from brain activity, positioning fMRI-to-image reconstruction as a promising bridge between neuroscience and computer vision. However, current methods predominantly rely on subject-specific models or require subject-specific fine-tuning, limiting their scalability and real-world applicability. In this work, we introduce \methodname{}, the first zero-shot brain visual decoding framework that eliminates the need for subject-specific adaptation. \methodname{} is built on the key insight that fMRI representations can be decomposed into subject-related and semantic-related components. By leveraging adversarial training, our method explicitly disentangles these components to isolate subject-invariant, semantic-specific representations. This disentanglement allows \methodname{} to generalize to unseen subjects without any additional fMRI data or retraining. Extensive experiments show that \methodname{} significantly outperforms zero-shot baselines and achieves performance comparable to fully finetuned models on several metrics. Our work represents a scalable and practical step toward universal neural decoding. Code and model weights are available at: \href{https://github.com/xmed-lab/ZEBRA}{{https://github.com/xmed-lab/ZEBRA}}.

\end{abstract}

%% file: sec/1_intro.tex
\section{Introduction}

The compelling connection between neural decoding and visual understanding has positioned fMRI-to-image reconstruction~\cite{takagi2023high,ozcelik2023natural,scotti2024mindeye,mindeye2,huo2024neuropictor,gong2025mindtuner} at the forefront of computational neuroscience and computer vision. As a non-invasive method for observing activity in the brain's visual cortex, fMRI signals offer the intriguing possibility of reverse-engineering human perception—translating blood-oxygen-level-dependent (BOLD) responses into detailed visual reconstructions of what a person sees. This ability, often described as a "brain camera," marks a major shift, with potential applications in mental state interpretation~\cite{MSI_thornton2024neural} and advanced brain-computer interfaces~\cite{BCI_sitaram2007fmri,BCI_sorger2012real}.

\begin{figure}
    \centering
    \includegraphics[width=1\linewidth]{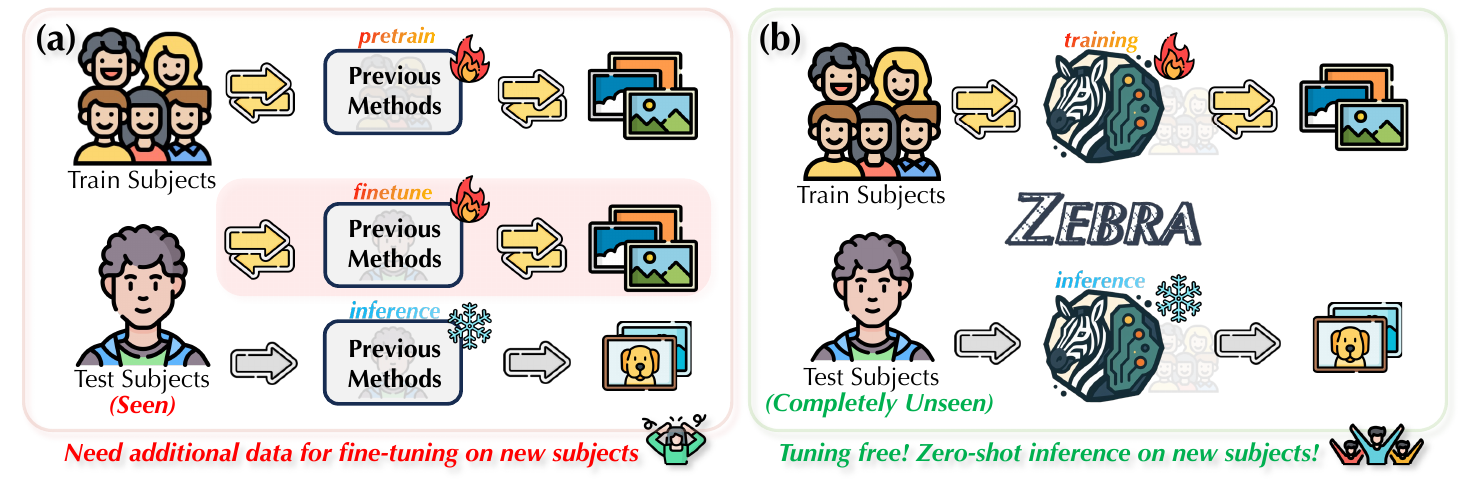}
    \caption{\textbf{(a)} Previous methods~\cite{mindeye2,huo2024neuropictor,gong2025mindtuner} typically involve two training stages: (1) pretraining a brain model with multiple subjects, and (2) fine-tuning the model for a specific subject. In this approach, the test subject is known to the model, which limits its zero-shot capability for new subjects. \textbf{(b)} In contrast, \methodname{} eliminates the fine-tuning stage, requiring training only once with the training subjects. This allows it to perform zero-shot inference on unseen subjects, achieving comparable performance to the fine-tuned approaches.}
    \label{fig:comparison}
\end{figure}

Despite remarkable progress in reconstructing images from individual brain data, the field faces a critical challenge: current models struggle to generalize across individuals. This limitation risks confining such breakthroughs to research labs rather than enabling real-world applications.
Recent efforts~\cite{mindeye2, gong2025mindtuner} have attempted to address this issue by developing cross-subject reconstruction through a two-step approach: first, \textit{pretraining} a unified model on multi-subject data, followed by subject-specific \textit{finetuning}, as illustrated in Fig.~\ref{fig:comparison}a. However, subject-specific finetuning imposes significant barriers to practical clinical use due to several limitations: (1)
Clinicians and neuroscientists must still depend on AI experts to fine-tune models for each new patient. (2)
The fine-tuning process is time-intensive, often taking around a day, which hampers real-time applications in brain-computer interfaces~\cite{BCI_sitaram2007fmri,BCI_sorger2012real} and neurorehabilitation~\cite{wang2018potential}. (3) There is no universal feature space capable of learning neural representations across human subjects, restricting broader exploration in cognitive functions and variability.
Thus, while developing \textbf{zero-shot cross-subject generalization methods} for brain visual decoding is critical, this area remains largely unexplored.

A straightforward approach would be to evaluate the previous state-of-the-art method, MindTuner~\cite{gong2025mindtuner}, in a zero-shot setting and attempt to improve upon it. However, this is infeasible due to its subject-specific design, which is tailored to certain information content of representations across subjects. As a result, it fails when tested on a new subject whose representation carries a different amount of information.
NeuroPictor~\cite{huo2024neuropictor} offers valuable insights by transforming fMRI data from different subjects into uniformly shaped 2D representations with spatial information preserved, facilitating the learning of a shared latent space.
Nonetheless, its zero-shot performance remains limited since it is sensitive to subject noise, as shown in Fig.~\ref{fig:umap}, and thus fails to learn invariant representations across subjects.
Motivated by these observations, one may consider combining NeuroPictor’s powerful unified brain encoding with the powerful decoding of MindTuner. Yet, even this possibly strongest baseline fails to achieve satisfactory results, as evidenced by the ``Our baseline'' and ``NeuroPictor$^\star$'' rows in Table~\ref{tab:main_res}.

\setlength\intextsep{0ex}
\begin{wrapfigure}{r}{0.45\textwidth}
\centering
    \includegraphics[width=1\linewidth]{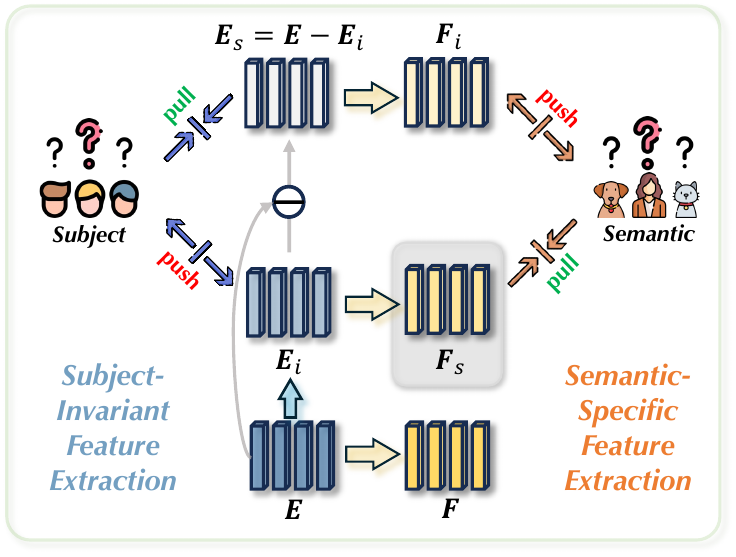}
    \caption{Core idea of \methodname{}. $\bm{F}_s$ is used as diffusion prior guidance.}
    \label{fig:core}
\vspace{1ex}
\end{wrapfigure}

To pave the way for zero-shot brain visual decoding, build on this baseline, we propose \methodname{}—\textbf{the first \underline{ze}ro-shot \underline{bra}in visual decoding framework}—that enables direct generalization to unseen subjects without requiring additional fMRI data or model retraining (Fig.~\ref{fig:comparison}b). 
As illustrated in Fig.~\ref{fig:core}, the core idea of \methodname{} is to disentangle fMRI-derived features into four complementary components, with a focus on learning \textit{subject-invariant} and \textit{semantic-specific} representations. This design is motivated by neuroscientific evidence that, despite inter-individual variability in brain activity, the human cortex encodes semantic information in a consistent and topographically organized manner across subjects~\cite{semantic_leeds2013comparing,semantic_lahnakoski2014synchronous,huth2016natural}.
For a reconstruction framework to generalize effectively across individuals, it should preserve \textbf{subject-invariant} (universal brain representations) and \textbf{semantic-specific} (class-discriminative) components, while suppressing subject-specific~\cite{specific_de2019individual,specific_moutsiana2016cortical} and semantically irrelevant variations.
To this end, \methodname{} first extracts subject-invariant features by removing subject-specific noise via residual decomposition and adversarial training. In parallel, semantic-specific features are projected into a shared visual-semantic space and aligned with CLIP embeddings, ensuring semantic-level discriminability while remaining agnostic to subject identity. This disentanglement strategy enables robust cross-subject generalization and facilitates scalable fMRI-to-image decoding.

Extensive experiments validate the effectiveness of \methodname{}, particularly on low-level perceptual and pixel-wise metrics. \methodname{} achieves a substantial improvement in PixCorr, with a gain of +0.084 (0.153 vs. 0.069 of NeuroPictor), and an average improvement of +6.4 percentage points (81.8\% vs. 75.4\%) on Alex (5).
More importantly, \methodname{} shows performance that is comparable to fully-finetuned methods in several metrics. For instance, it achieves an SSIM of 0.384, close to 0.375 of NeuroPictor (fully finetuned), despite not using any test subject data. 
Qualitative results further confirm that the visual reconstructions generated by \methodname{} are competitive with those produced by fully finetuned subject-specific models.

Our contributions can be summarized as follows: (1) We propose \methodname{}, the first zero-shot brain visual decoding framework that generalizes to unseen subjects without requiring additional fMRI data or finetuning; (2) We introduce a novel disentanglement strategy combining adversarial training and residual decomposition to learn subject-invariant and semantic-specific representations from fMRI signals; (3) We demonstrate that \methodname{} achieves competitive performance compared to few-shot and fully fine-tuned subject-specific methods across multiple quantitative and qualitative benchmarks.

%% file: sec/2_related_works.tex
\section{Related Works}

\subsection{fMRI-to-Image Reconstruction}
In recent years, with the rapid advancement and widespread adoption of functional magnetic resonance imaging (fMRI), researchers have increasingly recognized the critical value of fMRI signals for neuroscience research. Leveraging advanced technological tools, several datasets linking fMRI signals to images have emerged~\cite{NSD,BOLD,GOD}, with the Natural Scenes Dataset (NSD) being an exemplary instance~\cite{NSD}. Utilizing these datasets, the task of fMRI-to-image reconstruction has undergone significant development.
Early research predominantly employed neural networks such as Variational Autoencoders (VAEs) and Generative Adversarial Networks (GANs) for reconstructing images from fMRI signals~\cite{GAN1,GAN2,GAN3,GAN4}. However, these studies were either restricted to reconstructing simplistic digit images or lacked the capability to effectively represent both high-level and low-level semantic details clearly. The advent of diffusion models marked a transformative milestone, catalyzing numerous novel studies~\cite{mindeye2, gong2025mindtuner,fang2023alleviating, zengControllableMindVisual2024, scotti2023reconstructing, chenSeeingBrainConditional2023a, huo2024neuropictor, ferrante2023brain}. Diffusion-based methods excel in capturing complex, high-dimensional semantic information. To effectively represent low-level spatial details, some approaches have incorporated blurry images as control inputs~\cite{mindeye2,gong2025mindtuner}, while others have utilized contours or masks~\cite{zengControllableMindVisual2024,ferrante2023brain}. Moreover, certain studies have further refined high-dimensional semantic content by leveraging captions or labels~\cite{mindeye2,gong2025mindtuner,scotti2023reconstructing}. By integrating both high-level semantic guidance and low-level spatial control, many recent approaches have produced intuitive and compelling reconstruction outcomes.

\subsection{Cross-Subject fMRI-to-Image Reconstruction}
In the medical domain, fMRI signals exhibit substantial variability across individuals due to differences in anatomical and physiological structures. Given the challenges associated with acquiring fMRI data, research focusing on cross-subject adaptation is profoundly significant. Central to cross-subject studies is the alignment of fMRI signals across different subjects. While anatomical alignment provides a foundational step, numerous studies emphasize the greater importance of functional alignment~\cite{anatomy1,anatomy2}
Numerous studies tried to address this issue~\cite{wangMindBridgeCrossSubjectBrain2024, ferranteTheirEyesMultisubject2024, thualAligningBrainFunctions2023, xiaUMBRAEUnifiedMultimodal2024, qian2023fmri, li2024enhancing, scotti2024mindeye}.
Initial approaches typically trained separate models for each subject, resulting in poor generalizability and substantial data dependency issues~\cite{scotti2024mindeye}. Subsequent methodologies, such as MindBridge~\cite{wangMindBridgeCrossSubjectBrain2024} and MindEye2~\cite{mindeye2}, sought to develop subject-agnostic networks by constructing shared latent space. However, these models often require extensive fine-tuning when encountering new subjects. While MindTuner~\cite{gong2025mindtuner} partially mitigates this requirement, it does not fundamentally resolve the persistent reliance on subject-specific data in cross-subject fMRI research.

%% file: sec/3_method.tex
\begin{figure}[tb]
    \centering
    \includegraphics[width=1\linewidth]{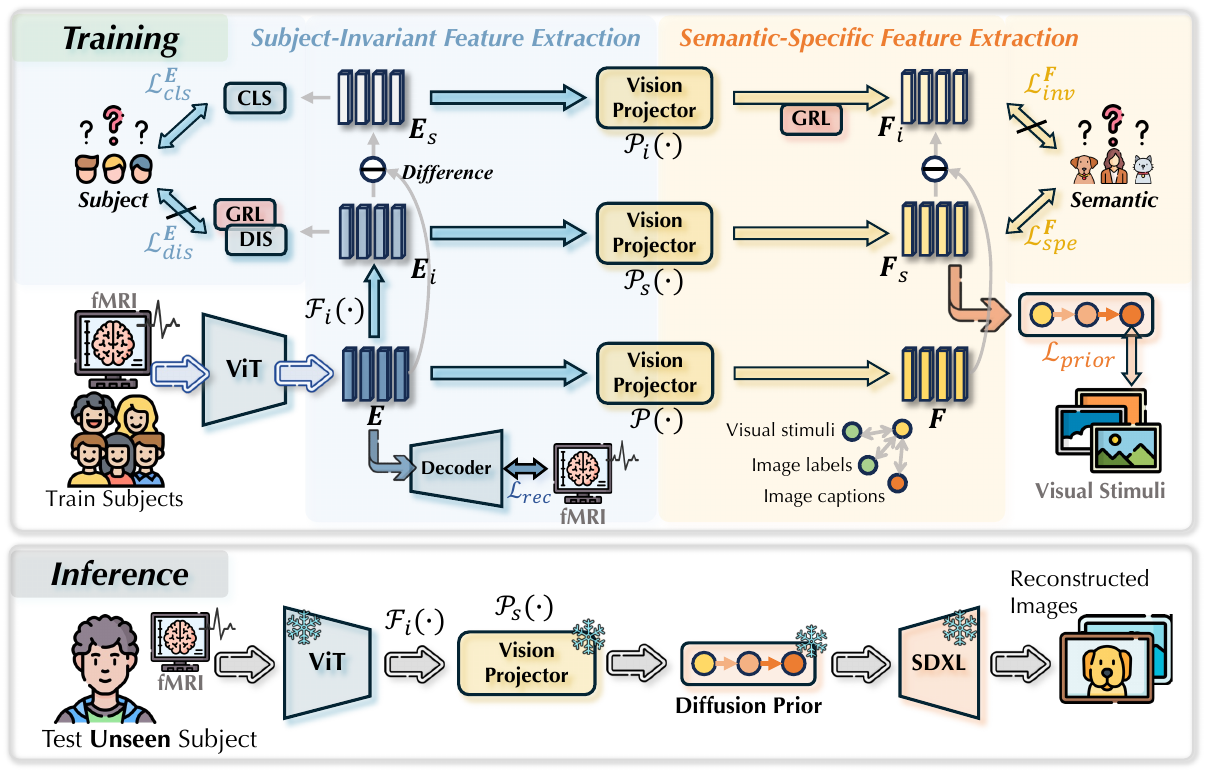}
    \caption{\methodname{} consists of two key components: (1) \textbf{Subject-Invariant Feature Extraction}, which disentangles subject-invariant representations from brain activity using adversarial learning and residual decomposition (\S\ref{sec:SIFE}); and (2) \textbf{Semantic-Specific Feature Extraction}, which aligns semantic information in brain features with vision-language embeddings via supervised learning and gradient reversal (\S\ref{sec:SSFE}). During inference, only the invariant projection path is used, enabling zero-shot generalization to unseen subjects.}
    \label{fig:framework}
\end{figure}

\section{Method}

As illustrated in Fig.~\ref{fig:framework}, based on a baseline with a ViT-based brain encoding backbone and a unCLIP generative model \S\ref{sec:baseline}, \methodname{} improves this with two main components: (1) a Subject-Invariant Feature Extraction module that maps brain visual representations to shared latent space by exploring subject-invariant features (\S \ref{sec:SIFE}), and (2) a Semantic-Specific Feature Extraction module that injects semantic features into the shared latent space. (\S \ref{sec:SSFE}).

\subsection{Baseline Framework}\label{sec:baseline}

Our baseline framework primarily consists of a ViT-based fMRI encoder (fMRI-PTE~\cite{qian2023fmri}, pretrained on the UK Biobank dataset~\cite{miller2016multimodal}) that maps fMRI data from different subjects into a shared latent space, and a diffusion prior network that converts the latent brain embeddings into vision features for image generation using Stable Diffusion.
Given an fMRI scan and its corresponding visual stimulus $y$, we first transform the fMRI data into a unified 2D brain activation map~\cite{qian2023fmri}, resulting in a single-channel image $x \in \mathbb{R}^{256 \times 256}$. The fMRI encoder then converts this 2D surface map into a latent representation $\bm{E} \in \mathbb{R}^{B \times L \times C_1}$, where $B$ is the batch size, $L$ is the number of tokens, and $C_1$ is the brain feature dimension.
Then the latent representation $\bm{E}$ is converted to embeddings in CLIP space $\bm{F} \in \mathbb{R}^{B \times L \times C_2}$ as guidance for reconstruction. 
Following MindEye2~\cite{mindeye2}, we utilize a diffusion prior~\cite{ramesh2022hierarchical} to transform the fMRI-CLIP embedding $\bm{F}$ into a reconstructed OpenCLIP vision embedding $\bm{F}_y$ of the corresponding visual stimulus. Similar to DALL·E 2~\cite{ramesh2022hierarchical}, the diffusion prior is trained to minimize the mean squared error (MSE) between predicted and target embeddings:
\begin{equation}
    \mathcal{L}_{prior} = \mathbb{E}_{\bm{F}_{y}, \bm{F} \epsilon \sim \mathcal{N}(0, 1)} 
    \left\| \epsilon(\bm{F}) - \bm{F}_y \right\|^2.
\end{equation}
The training of the baseline model involves three losses: 
(1) a contrastive loss $\mathcal{L}_\mathrm{CLIP_t}$ between the predicted CLIP text embedding $\bm{F}^t$ and the ground truth $\bm{F}^t_y$; 
(2) a contrastive loss $\mathcal{L}_\mathrm{CLIP_v}$ between the predicted CLIP vision embedding $\bm{F} \in \mathbb{R}^{B \times N \times C}$ and the ground truth $\bm{F}^v_y$; and 
(3) the diffusion prior loss $\mathcal{L}_{prior}$. 
Both $\mathcal{L}_\mathrm{CLIP_t}$ and $\mathcal{L}_\mathrm{CLIP_y}$ adopt the BiMixCo loss, which aligns video frames and corresponding fMRI signals using a bidirectional contrastive objective and MixCo-based data augmentation, detailed in the Supplementary.

\subsection{Subject-Invariant Feature Extraction}\label{sec:SIFE}
\textbf{Residual Decomposition \& Adversarial Training.}
fMRI signals are highly idiosyncratic across individuals, making direct modeling challenging. To enable generalization, it is essential to filter out subject-specific noise and retain only invariant, stimulus-relevant components.
The goal of \textit{Subject-Invariant Feature Extraction (SIFE)} is to disentangle general brain representations into two components: subject-invariant features and subject-specific features. This disentanglement is achieved via \textit{residual decomposition}, where we employ self-attention blocks $\mathcal{F}_i(\cdot)$ as the invariant feature extractor to derive subject-invariant features: $\bm{E}_i = \mathcal{F}_i(\bm{E})$. 
The choice of self-attention is not essential to enforcing invariance—which is primarily driven by the gradient reversal layer (GRL) and the associated adversarial losses—but serves to maintain architectural consistency with the brain encoder based on ViT. 
Subsequently, the subject-specific features are obtained as the residual difference: $\bm{E}_s = \bm{E} - \bm{E}_i$.
This residual decomposition ensures that the extracted $\bm{E}_i$ captures components common across subjects, while $\bm{E}_s$ accounts for individual variability. The decomposition is further regularized by adversarial and supervised objectives to enforce disentanglement constraints and semantic consistency.
To ensure that $\bm{E}_i$ is truly invariant to subject identity, we apply an adversarial training strategy~\cite{DG_li2018deep}. Specifically, a subject discriminator $\mathcal{D}_{dis}$ is trained to maximize its ability to predict the subject label from $\bm{E}_i$, while the invariant extractor $\mathcal{F}_i$ aims to produce features that prevent $\mathcal{D}_{dis}$ from correctly identifying the subject. This adversarial objective is formulated as the following min-max game:
\begin{equation}\label{eq:L_E_dis}
\underset{\theta_{\mathcal{E}},\theta_{\mathcal{F}}}{\min} \text{ } \underset{\theta_{\mathcal{D}_{dis}}}{\max}\left \{ \mathcal{L}_{dis}^{\bm{E}} := -\mathbb{E}_{x,s\sim \mathcal{X},\mathcal{S}}\left [s\log \mathcal{D}_{dis}(\mathcal{E}(\mathcal{F}_i(\bm{E})) \right ] \right \},
\end{equation}
where $s$ denotes the subject label.
To guide the learning of subject-specific features $\bm{E}_s$, we introduce a subject classifier $\mathcal{D}_{cls}$, which is trained to predict the subject identity from $\bm{E}_s$. Its parameters $\theta_{\mathcal{D}_{cls}}$ are optimized via the following classification loss:
\begin{equation}
    \label{eq:L_E_cls}
    \underset{\theta_{\mathcal{D}_{cls}}}{\min} \left \{ \mathcal{L}_{cls}^{\bm{E}} := -\mathbb{E}_{x,s\sim \mathcal{X},\mathcal{S}}\left [s\log \mathcal{D}_{cls}(\bm{E}_s) \right ]  \right \}.
\end{equation}
The residual decomposition mechanism inherently reduces the subject-specific signal in $\bm{E}_i$, thereby promoting the inclusion of subject-invariant content. We assume that subject-irrelevant features largely overlap with shared semantic representations. Thus, enforcing $\bm{E}_i$ to be adversarially invariant ensures that it encodes generalizable brain features across individuals.

\textbf{Representation Preservation Anchor.} 
While adversarial training promotes subject-invariant representation learning, it may inadvertently distort the original high-dimensional brain feature space $\bm{E}$. To counteract this, we introduce a \textit{representation preservation anchor} via an auxiliary fMRI reconstruction task. Specifically, we employ a masked decoder $\mathcal{D}_{rec}(\cdot)$—comprising two deconvolution layers followed by a linear prediction head—to reconstruct the input signal as $\hat{x} = \mathcal{D}_{rec}(\bm{E})$.
We adopt the mean absolute error (MAE) as the reconstruction loss, defined as:
\begin{equation}
\label{eq:rec_loss}
\mathcal{L}_{rec} = \mathbb{E}_{(x, \hat{x}) \sim \mathcal{X}} \left[ \left| \hat{x} - x \right| \right].
\end{equation}
This reconstruction task serves as an anchor to preserve essential neural information in the latent space, ensuring $\bm{E}$ retains both biological fidelity and semantic coherence under adversarial training.

\subsection{Semantic-Specific Feature Extraction}\label{sec:SSFE}

Given the subject-invariant brain representation $\bm{E}_i$ from SIFE, we further inject semantic information from stimuli to enhance semantic-specific alignment. Similar to SIFE, the \textit{Semantic-Specific Feature Extraction (SSFE)} module consists of two components: Adversarial Training and a Representation Preservation Anchor.

\textbf{Adversarial Training.}
We project brain features into the CLIP vision space using vision projectors composed of three linear layers with GELU activation~\cite{GELU}, yielding three types of CLIP-aligned embeddings: semantic-specific features $\bm{F}_s = \mathcal{P}_s(\bm{E}_i)$, semantic-invariant features $\bm{F}_i = \mathcal{P}_i(\bm{E}_s)$, and general representations $\bm{F} = \mathcal{P}(\bm{E})$. The disentanglement between $\bm{F}_s$ and $\bm{F}_i$ is driven by the residual decomposition in the brain feature space.
To ensure $\bm{F}_s$ captures meaningful semantic information, we directly align it with OpenCLIP vision embeddings $\bm{F}_y$, using the BiMixCo loss $\mathcal{L}_{spe}^{\bm{F}}$, detailed in the Supplementary.

\setlength\intextsep{0ex}
\begin{wrapfigure}{r}{0.6\textwidth}
\centering
    \includegraphics[width=1\linewidth]{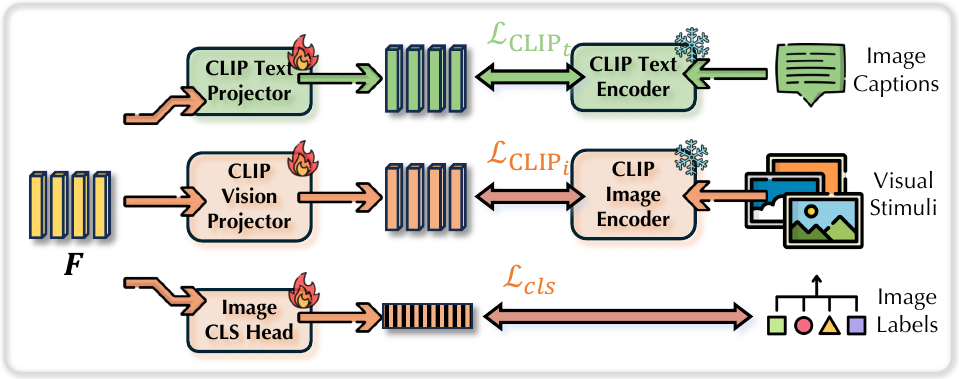}
    \caption{Representation Preservation Anchor of SSFE.}
    \label{fig:anchor}
\vspace{1ex}
\end{wrapfigure}
To reinforce the semantic purity of $\bm{F}_s$, we apply adversarial training to $\bm{F}_i$, encouraging it to encode minimal semantic content. As a result of the residual structure, this pushes more semantic information into $\bm{F}_s$. Specifically, we prepend a gradient reversal layer~\cite{GRL} before the projection: $\bm{F}_i = \mathcal{P}_i[\text{GRL}(\bm{E}_s)]$, discouraging $\bm{E}_s$ from aligning with CLIP target features. The corresponding adversarial loss $\mathcal{L}_{inv}^{\bm{F}}$ is also a BiMixCo loss with min-max game similar to Eq.~\eqref{eq:L_E_dis} and omitted here.

\textbf{Representation Preservation Anchor.} 
Similar to SIFE, to ensure semantic consistency in the latent space, we introduce a preservation anchor by aligning CLIP embeddings across three perspectives: classification, vision, and text. We employ a linear classifier $\mathcal{C}(\cdot)$ on the $\bm{E}$ to predict the image label $\hat{c}$, with a cross-entropy loss $\mathcal{L}_{\text{cls}}$. For vision and text alignments, we adopt BiMixCo losses $\mathcal{L}_{\text{CLIP}_v}$ and $\mathcal{L}_{\text{CLIP}_t}$ to align to CLIP image and text embeddings. The total semantic loss is $\mathcal{L}_{\text{sem}} = \mathcal{L}_{\text{cls}} + \mathcal{L}_{\text{CLIP}_v} + \mathcal{L}_{\text{CLIP}_t}$.
The overall training loss can be described as:
\begin{equation}
    \label{semantic-reconstruction-loss} \mathcal{L}=\mathcal{L}_{rec} + \mathcal{L}^{\bm{E}}_{dis} + \mathcal{L}^{\bm{E}}_{cls} + \mathcal{L}^{\bm{F}}_{inv} + \mathcal{L}^{\bm{F}}_{spe} + \mathcal{L}_{sem} +\lambda\mathcal{L}_{prior}, 
\end{equation}
where $\lambda$ is set to 30 following previous methods~\cite{mindeye2}.

%% file: sec/4_exp.tex
\section{Experiments}

\subsection{Experimental Setup}
\noindent
{\bf Dataset.} 
We use the Natural Scenes Dataset (NSD) \cite{NSD} for both training and evaluation. NSD contains visual image stimulus and corresponding fMRI recordings of 8 subjects, with each subject viewing 8,000-9,000 images. The original images are collected from MS-COCO dataset \cite{lin2014microsoft}, which are consisted of complex natural images. Following \cite{ozcelik2023natural}, we use the corresponding captions of the images in COCO dataset for training. For both training and evaluation, we average three trials of fMRI signal of the same images following \cite{huo2024neuropictor}. 
The final results were tested on subjects 1, 2, 5 or 7, since these subjects complete all scanning sessions, sharing the same $982$ images as testing data. For each test subject, we use all other 7 subjects to train the model and tested on the unseen subject with unseen test split.

\noindent
{\bf Evaluation Metrics.} 
We follow the metrics of Mindeye2~\cite{scotti2024mindeye} to evaluate both high-level and low-level consistency. On the low-level aspect, we use pixelwise correlation, Structural Similarity Index Metric (SSIM)~\cite{ssim}, AlexNet(2), and AlexNet(5). High-level metrics are calculated by extracting features using specific networks, including EffNet-B~\cite{tan2019efficientnet}, SwAV~\cite{swav}, Inception~\cite{inception}, and CLIP~\cite{radford2021clip}. Please refer to the Supplementary for more details.

\noindent
\textbf{Implementation Details.} 
All experiments were conducted for 60 epochs using 8 NVIDIA RTX H800 GPUs with a total batch size of 128 (16 samples per GPU). We adopt the AdamW optimizer~\cite{loshchilov2017adamw} with a learning rate of 1e-4, following the OneCycle learning rate schedule~\cite{smith2019super}.
In the inference stage, we follow MindEye2's two-stage decoding process. First, the predicted image latents are decoded into coarse images using SDXL unCLIP. These coarse outputs are then refined using base SDXL in image-to-image mode, guided by predicted captions. The refinement starts from a noised version of the coarse image, skipping the first 50\% of diffusion steps.

\begin{table}[!t]
  \centering
  \caption{
Quantitative comparison of \methodname{} against representative methods under different training regimes. Results are averaged over subjects 1, 2, 5, and 7 from the Natural Scenes Dataset. \textit{Fully fine-tuned} methods are trained on 40-hour data from the test subject, \textit{few-shot} methods use only 1-hour recordings, while \textit{zero-shot} methods do not use any data from the test subjects. “NeuroPictor$^\dagger$” denotes a version pretrained on 40-hour data from all subjects without fine-tuning on the test subjects. “NeuroPictor$^\star$” represents our implementation in a zero-shot setting (pretrained on the other 7 subjects). 
Previous methods—excluding NeuroPictor—are fundamentally infeasible in zero-shot scenarios because their architectures depend on subject-specific linear mappings.
``Our baseline'' refers to the strong baseline (\S\ref{sec:baseline}) combining NeuroPictor and MindTuner. 
}
  \resizebox{\linewidth}{!}{
  \begin{tabular}{lcccccccc}
    \toprule
    \multirow{2}{*}{Method} & \multicolumn{4}{c}{Low-Level}& \multicolumn{4}{c}{High-Level}\\
    \cmidrule(lr){2-5} \cmidrule(lr){6-9} 
     &PixCorr$\uparrow$&SSIM$\uparrow$&Alex(2)$\uparrow$&Alex(5)$\uparrow$&Incep$\uparrow$&CLIP$\uparrow$&Eff$\downarrow$&SwAV$\downarrow$\\ \midrule 

     \rowcolor{gray!20}\multicolumn{9}{c}{\textit{Fully fine-tuned}} \\  

    \midrule
    Takagi...~\cite{takagi2023high} \scriptsize\textcolor{gray}{[CVPR'23]}                     & 0.246  & {0.410}  & {78.9\%} & {85.6\%} & 83.8\% & 82.1\% & {0.811}  & {0.504}    \\

    Ozcelik...~\cite{ozcelik2023natural} \scriptsize\textcolor{gray}{[Sci. Rep.'23]}                  & 0.273  & {0.365}  & 94.4\% & 96.6\% & 91.3\% & 90.9\% & 0.728  & 0.422  \\
    
    MindEye1~\cite{scotti2024mindeye} \scriptsize\textcolor{gray}{[NeurIPS'24]}                     & 0.319  & {0.360}  & 92.8\% & 96.9\% & 94.6\% & 93.3\% & 0.648  & 0.377  \\
    UMBRAE~\cite{xiaUMBRAEUnifiedMultimodal2024} \scriptsize\textcolor{gray}{[ECCV'24]}
        & 0.283 & {0.341} & 95.5\% & 97.0\% & 91.7\% & 93.5\% & 0.700 & 0.393 \\

    NeuroPictor~\cite{huo2024neuropictor} \scriptsize\textcolor{gray}{[ECCV'24]}
        & 0.229 & {0.375} & 96.5\% & 98.4\% & 94.5\% & 93.3\% & 0.639 & 0.350  \\
        
    NeuroPictor$^\dagger$\cite{huo2024neuropictor} \scriptsize\textcolor{gray}{[ECCV'24]}
    & {0.141} & {0.349} & 91.4\% & 95.7\% & 88.3\% & 88.9\% & 0.722 & 0.417  \\

    MindBridge~\cite{wangMindBridgeCrossSubjectBrain2024} \scriptsize\textcolor{gray}{[CVPR'24]}                    & {0.151}  & {0.263}  & 87.7\% & 95.5\% & 92.4\% & 94.7\% & 0.712  & 0.418 \\
        
    MindEye2~\cite{mindeye2} \scriptsize\textcolor{gray}{[ICML'24]}                    & 0.322  & 0.431  & 96.1\% & 98.6\% & 95.4\% & 93.0\% & 0.619  & 0.344  \\
    
    MindTuner~\cite{gong2025mindtuner} \scriptsize\textcolor{gray}{[AAAI'25]}           & 0.322  & 0.421  & 95.8\% & 98.8\% & 95.6\% & 93.8\% & 0.612  & 0.340 \\
    \midrule 
    \rowcolor{gray!20}\multicolumn{9}{c}{\textit{Few-shot}} \\  \midrule

    MindEye2~\cite{mindeye2} (1 hour)           & 0.195  & 0.419  & 84.2\% & 90.6\% & 81.2\% & 79.2\% & {0.810}  & {0.468}  \\
    
    MindTuner~\cite{gong2025mindtuner} (1 hour)          & 0.224  & 0.420  & 87.8\% & 93.6\% & 84.8\% & 83.5\% & {0.780}  & {0.440} \\

    \midrule 
    \rowcolor{gray!20}\multicolumn{9}{c}{\textit{Zero-shot}} \\  \midrule 

    NeuroPictor$^\star$  & 0.057  & 0.297  & 71.4\% & 74.7\% & 62.5\% & 66.0\% & 0.939  & 0.607\\
    Our baseline         & 0.074  & 0.316  & 70.8\% & 74.0\% & 63.5\% & 62.5\% & 0.920  & 0.602\\
    \rowcolor{cyan!5}\methodname{} & 0.131  & 0.375  & 74.6\% & 81.2\% & 72.2\% & 71.5\% & 0.837  & 0.506\\
    
     \bottomrule
  \end{tabular}}
  \label{tab:main_res}
\end{table}

\begin{figure}[ht]
    \centering
\includegraphics[width=1\linewidth]{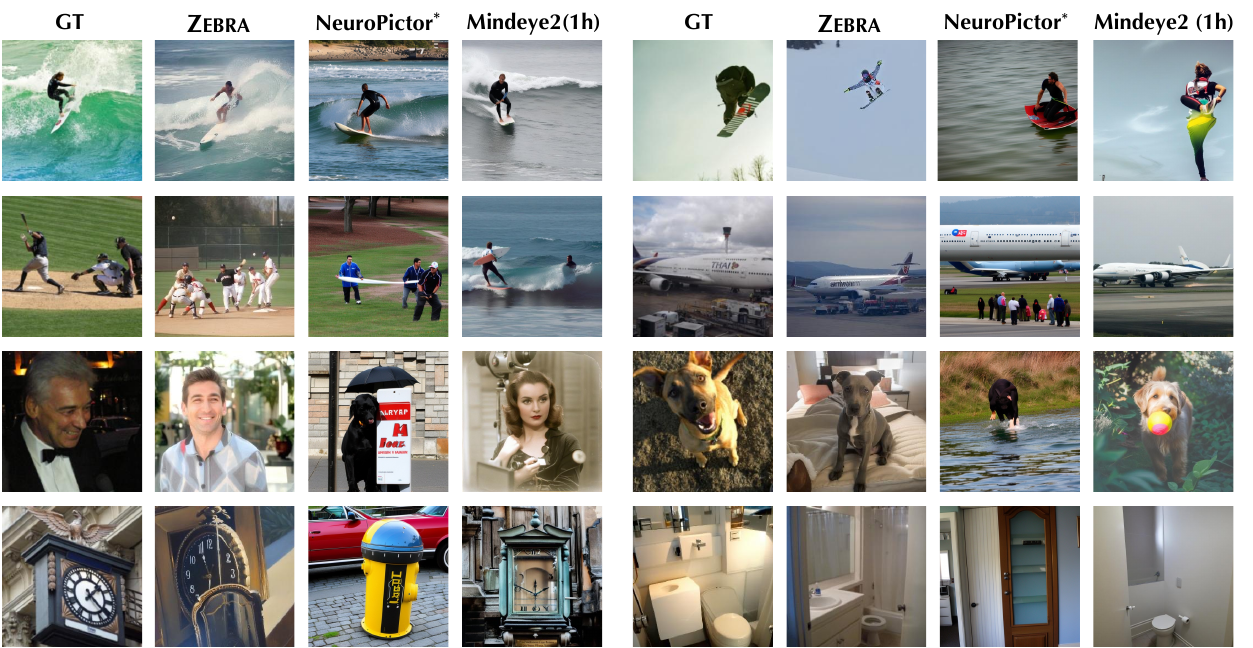}
    \caption{Qualitative comparison between \methodname{} and zero-shot implementation of NeuroPictor and Mindeye2 (1h).}
    \label{fig:SOTA_vis}
\end{figure}

\begin{table}[!t]
  \centering
  \caption{Ablations on the key components of \methodname{}, and all results are from subject 1. `Adv.' denotes adversarial training and `Anchor' stands for preservation anchor.}
  
  \resizebox{\linewidth}{!}{
  \begin{tabular}{@{}lcccc|cccccccc@{}}
    \toprule
    Base & \multicolumn{2}{c}{SIFE} & \multicolumn{2}{c|}{SSFE} & \multicolumn{4}{c}{Low-Level} & \multicolumn{4}{c}{High-Level} \\
    \cmidrule(lr){2-3} \cmidrule(lr){4-5} \cmidrule(lr){6-9} \cmidrule(lr){10-13}
    line & Adv. & Anchor & Adv. & Anchor  & PixCorr$\uparrow$ & SSIM$\uparrow$ & Alex(2)$\uparrow$ & Alex(5)$\uparrow$ & Incep$\uparrow$ & CLIP$\uparrow$ & Eff$\downarrow$ & SwAV$\downarrow$  \\
    \midrule

    \ding{51} &        &        &        &        & 0.089 & 0.325 & 72.5\% & 74.7\% & 64.7\% & 63.2\% & 0.891 & 0.579 \\
    
    \ding{51} & \ding{51} &        &        &        & 0.129 & 0.355 & 73.9\% & 77.4\% & 68.0\% & 66.8\% & 0.885 & 0.545 \\
     
    \ding{51} & \ding{51} & \ding{51} &        &        & 0.134 & 0.368 & 74.3\% & 78.3\% & 70.0\% & 69.3\% & 0.855 & 0.525 \\
    
    \ding{51} & \ding{51} & \ding{51} & \ding{51} &        & 0.142 & 0.374 & 75.2\% & 79.6\% & 71.4\% & 70.8\% & 0.832 & 0.505 \\
    
    \midrule
    \rowcolor{cyan!5}
    \ding{51} & \ding{51} & \ding{51} & \ding{51} & \ding{51} & 0.153 & 0.384 & 76.1\% & 81.8\% & 73.4\% & 72.3\% & 0.814 & 0.490 \\
    \bottomrule
  \end{tabular}}
  \label{tab:ablation}
\end{table}

\subsection{Main Results}
\textbf{Quantitative Results.} We evaluate \methodname{} against representative methods across various training regimes on the Natural Scenes Dataset, with results averaged over subjects 1, 2, 5, and 7. As shown in Table~\ref{tab:main_res}, \methodname{} achieves competitive performance without using any subject-specific data, highlighting its strong generalization ability in the \textit{zero-shot} setting.
Compared to the only other zero-shot-compatible baseline, NeuroPictor$^\star$, \methodname{} achieves substantial improvements across all metrics. On low-level similarity metrics, \methodname{} improves PixCorr from 0.057 to 0.131 and SSIM from 0.297 to 0.375. Similarly, high-level perceptual metrics show consistent gains: \methodname{} achieves 74.6\% and 81.2\% on AlexNet(2) and AlexNet(5), respectively, compared to 71.4\% and 74.7\% for NeuroPictor$^\star$.
On high-level semantic metrics, \methodname{} outperforms NeuroPictor$^\star$ with margins of +9.7\% on Inception (72.2\% vs. 62.5\%) and +5.5\% on CLIP similarity (71.5\% vs. 66.0\%). \methodname{} also demonstrates lower perceptual distance, with Eff decreasing from 0.939 to 0.837, and SwAV from 0.607 to 0.506, indicating stronger alignment with ground-truth representations.
These results demonstrate \methodname{}’s clear advantage in generalizing across subjects without requiring fine-tuning. While fully fine-tuned methods unsurprisingly perform better due to access to the data of test subjects, \methodname{} narrows this gap significantly. For instance, in the zero-shot setting, \methodname{} achieves 74.6\% on AlexNet(2), compared to 78.9\% by Takagi et al.~\cite{takagi2023high} and 87.7\% by MindBridge~\cite{wangMindBridgeCrossSubjectBrain2024}, despite using no data from the test subjects.
In summary, \methodname{} sets a new state of the art in zero-shot neural decoding, outperforming prior zero-shot methods by large margins across all metrics, and approaching the performance of few-shot and even some fully fine-tuned approaches.

\textbf{Qualitative Results.}
We compare \methodname{} with the zero-shot implementation of NeuroPictor$^\star$ and the few-shot baseline MindEye2 (1-hour) as shown in Fig.~\ref{fig:SOTA_vis}. 
Qualitatively, \methodname{} generates high-fidelity images that are visually comparable to those produced by fully supervised models, and clearly surpasses {NeuroPictor}$^\star$ in overall perceptual quality. 
Compared to few-shot methods, the main limitation of \methodname{} lies in semantic accuracy. As highlighted in the failure cases Fig.~\ref{fig:failure_cases}, \methodname{} tends to struggle with fine-grained semantic distinctions, especially for rare object categories.

\subsection{Ablation Studies}

\setlength\intextsep{0ex}
\begin{wrapfigure}{r}{0.56\textwidth}
\centering
    \includegraphics[width=1\linewidth]{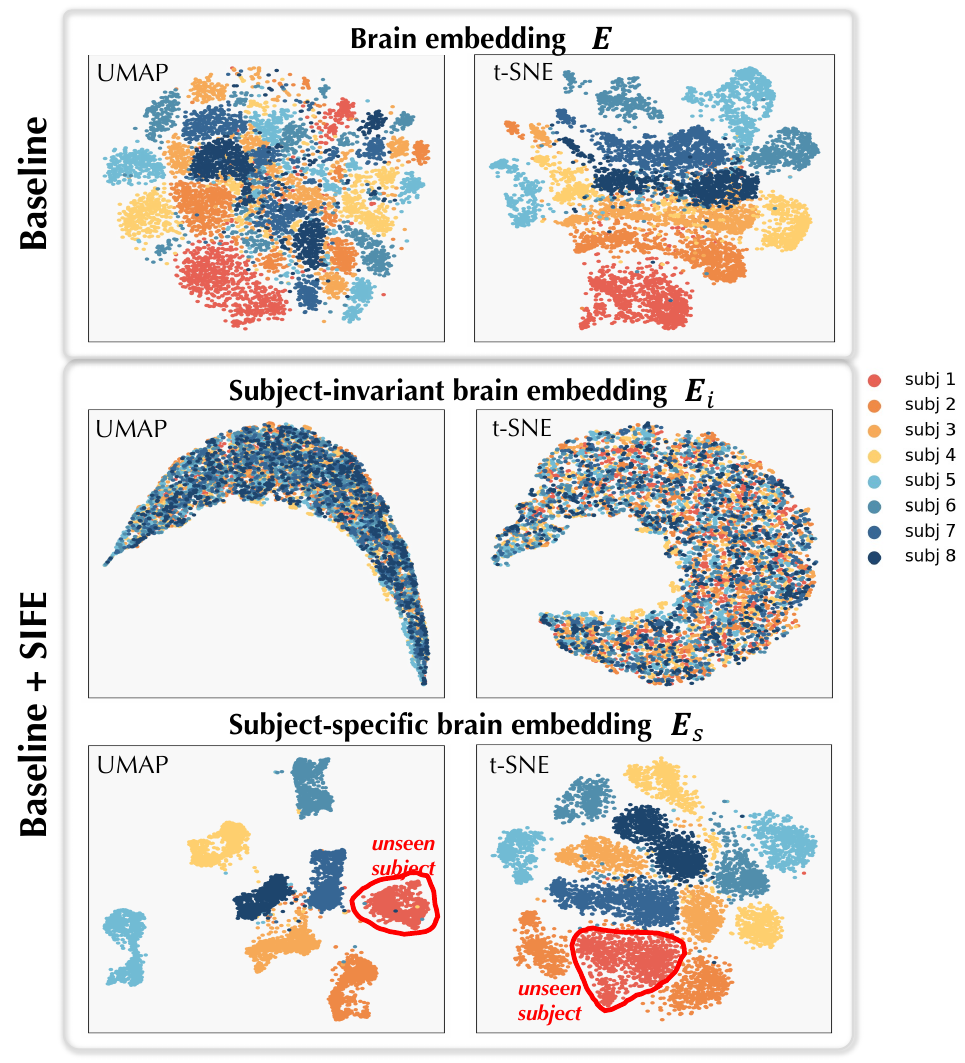}
    \caption{Visualization of subject-invariant and -specific features with UMAP and t-SNE. 
}
    \label{fig:umap}
\vspace{-2ex}
\end{wrapfigure}
We conduct ablation studies on Subject 1 (trained on Subjects 2-8) to assess the contribution of each component in \methodname{}, including adversarial training and the proposed preservation anchor in both the Subject-Invariant Feature Extractor (SIFE) and the Subject-Specific Feature Enhancer (SSFE).

\textbf{Effectiveness of the key components.}  
As shown in Table~\ref{tab:ablation}, the base model without any regularization performs poorly (e.g., PixCorr = 0.089, Alex(5) = 72.7\%). Adding adversarial training to SIFE yields clear improvements across both low- and high-level metrics (e.g., +0.040 PixCorr, +3.6\% CLIP). Adding the anchor further boosts performance, especially in high-level features like CLIP. Incorporating adversarial loss in SSFE brings additional gains, especially in high-level metrics, confirming the benefit of regularizing both branches. The full model with all components achieves the best overall results, highlighting the effectiveness of our design.

\textbf{Effectiveness of the proposed subject-invariant representation learning.} 
In Fig.~\ref{fig:umap}, we use UMAP and t-SNE to visualize the feature distributions of different subjects. 
The left two plots show the subject-invariant features $\bm{E}_i$, while the right two show the subject-specific features $\bm{E}_s$. 
In the case of $\bm{E}_i$, data points from all subjects are highly mixed without forming subject-specific clusters, demonstrating that the learned features are invariant to subject identity. 
In contrast, $\bm{E}_s$ exhibits clear clustering by subject, indicating that the model successfully captures individual-specific information. 
These results confirm that our method effectively disentangles subject-invariant and subject-specific representations.

\begin{table}[!t]
  \centering
  \caption{Ablation on the number of training subjects for \methodname{}, evaluated on data from subject 1.}
  \resizebox{0.75\linewidth}{!}{
  \begin{tabular}{@{}c|cccccccc@{}}
    \toprule
    \# of & \multicolumn{4}{c}{Low-Level} & \multicolumn{4}{c}{High-Level} \\
    \cmidrule(lr){2-5} \cmidrule(lr){6-9}
    Subjects & PixCorr$\uparrow$ & SSIM$\uparrow$ & Alex(2)$\uparrow$ & Alex(5)$\uparrow$ & Incep$\uparrow$ & CLIP$\uparrow$ & Eff$\downarrow$ & SwAV$\downarrow$ \\
    \midrule

    4 (2 to 5) & 0.109 & 0.325 & 68.2\% & 73.5\% & 66.1\% & 63.7\% & 0.871 & 0.563 \\
    
    5 (2 to 6) & 0.126 & 0.347 & 71.3\% & 76.8\% & 68.7\% & 67.2\% & 0.846 & 0.534 \\
     
    6 (2 to 7) & 0.135 & 0.363 & 74.1\% & 79.0\% & 70.8\% & 69.6\% & 0.823 & 0.508 \\
    
    \midrule
    \rowcolor{cyan!5}
    7 (2 to 8)  & \textbf{0.153} & \textbf{0.384} & \textbf{76.1\%} & \textbf{81.8\%} & \textbf{73.4\%} & \textbf{72.3\%} & \textbf{0.814} & \textbf{0.490} \\
    \bottomrule
  \end{tabular}}
  \label{tab:subj_num}
\end{table}

\textbf{Ablation on Number of Training Subjects.} 
As shown in Table~\ref{tab:subj_num}, we evaluate the impact of training subject count on model performance. As the number of subjects increases from 4 to 7, we observe consistent performance improvements across all metrics. For instance, PixCorr improves significantly from 0.109 to 0.153, and both low- and high-level semantic scores show similar trends, such as CLIP score increasing from 63.7\% to 72.3\%. These results demonstrate that incorporating more diverse subject data helps the model generalize better, highlighting its scalability and robustness.

%% file: sec/5_conclusion.tex
\section{Discussion and Conclusion}

\noindent\textbf{Key Contributions and Insights.} 
The novelty of our work lies not in individual architectural components, but in the problem formulation, representation disentanglement design, and cross-subject transfer mechanism that together enable zero-shot fMRI-to-image decoding. 
\underline{First, the importance of the zero-shot setting:} traditional fMRI decoding frameworks typically rely on fully supervised, subject-specific training, requiring separate model tuning for each individual under expert supervision. Such procedures are time-consuming—often exceeding 12 hours per subject—and computationally prohibitive for clinical use. In contrast, our approach introduces, for the first time, a zero-shot setting that allows the model to be directly applied to unseen subjects without retraining. This paradigm shift makes neural decoding fast, scalable, and clinically practical, achieving 73.4\% decoding performance with approximately one second of inference per image. 
\underline{Second, neuroscience-inspired representation disentanglement:} our design is motivated by neuroscientific evidence that, despite inter-individual variability, the human cortex encodes semantic information in a consistent and topographically organized manner across subjects. To preserve this universality while maintaining discriminative power, we explicitly separate subject-invariant and semantic-specific representations through the SIFE and SSFE modules, balancing fidelity and generalizability. 
\underline{Third, adversarial disentanglement with preservation anchors:} adversarial training objectives are employed to automatically extract invariant and specific features, while a Representation Preservation Anchor ensures that essential individual information is retained during zero-shot transfer. 
\underline{Finally, empirical superiority in zero-shot decoding:} our framework outperforms state-of-the-art baselines adapted to the zero-shot setting (e.g., MindTuner), underscoring the effectiveness of our disentanglement and transfer design in achieving robust cross-subject generalization.

\begin{figure}[t] 
  \centering
  \includegraphics[width=0.9\textwidth]{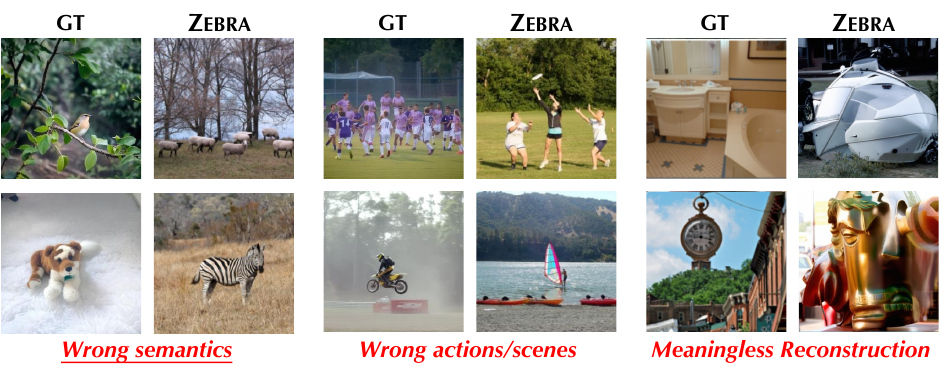}
  \caption{Failure cases, mainly caused by wrong semantics.}
  \label{fig:failure_cases}
\end{figure}

\textbf{Limitations and Future Work.}
Despite the encouraging results, several limitations remain. Although our reconstructed images exhibit competitive quality, especially in low-level perceptual metrics, their semantic fidelity still lags behind few-shot approaches. Improving high-level semantic accuracy remains a key challenge. Nonetheless, our work offers a promising zero-shot strategy and lays the foundation for building generalizable brain decoding models.
Another limitation lies in the scope of downstream tasks. While this study focuses on image reconstruction, the proposed \methodname{} framework is inherently modality-agnostic and could be extended to more complex domains such as text or video. For instance, integrating \methodname{} with existing methods like NeuroClips~\cite{gong2025neuroclips} or \textsc{Neurons}~\cite{wang2025neurons} could enable zero-shot fMRI-to-video generation, facilitating a richer understanding of human perceptual experiences.
Moreover, the current dataset contains a limited number of subjects, which restricts the ability to fully demonstrate the generalizability of our approach. We believe that as more training subjects and fMRI data become available, the model's robustness and zero-shot performance will further improve. Expanding subject diversity is especially important for advancing toward universal brain decoders.
Finally, additional fMRI recordings and broader subject coverage are essential to capture real-world visual experiences more comprehensively. Addressing these limitations will require interdisciplinary progress across machine learning, computer vision, neuroscience, and biomedical engineering. We emphasize that alongside technical advances, it is equally important to establish ethical and regulatory frameworks to ensure the privacy and responsible use of brain data.

\textbf{Conclusion.}
In this work, we introduced \methodname{}, a novel zero-shot brain visual decoding framework that addresses the critical challenge of generalizing fMRI-to-image reconstruction to unseen subjects. By disentangling subject-specific and semantic-specific components in the fMRI embedding space, \methodname{} enables accurate visual reconstruction without requiring additional data or retraining for new individuals. Our approach leverages adversarial learning and residual decomposition to isolate shared semantic representations, achieving strong generalization across subjects. Extensive experiments demonstrate that \methodname{} outperforms existing zero-shot baselines and approaches the performance of fully finetuned models, both quantitatively and qualitatively. This represents a significant step toward practical and scalable brain decoding systems with real-world applicability in neuroscience, clinical settings, and brain-computer interfaces.

\section*{Acknowledgements}
This work was partially supported by a grant from the Joint Research Scheme (JRS) under the National Natural Science Foundation of China (NSFC) and the Research Grants Council (RGC) of Hong Kong (Project No. N\_HKUST654/24), a grant from the RGC of the Hong Kong Special Administrative Region, China (Project No. R6005-24), and a grant from the RGC of the Hong Kong Special Administrative Region, China (Project No. AoE/E-601/24-N).

%% file: sec/checklist.tex
\newpage
\section*{NeurIPS Paper Checklist}



\begin{enumerate}

\item {\bf Claims}
    \item[] Question: Do the main claims made in the abstract and introduction accurately reflect the paper's contributions and scope?
    \item[] Answer: \answerYes{} 
    \item[] Justification: Main contributions and scope have been accurately claimed in the abstract and introduction.

\item {\bf Limitations}
    \item[] Question: Does the paper discuss the limitations of the work performed by the authors?
    \item[] Answer: \answerYes{} 
    \item[] Justification:  The limitations are detailedly discussed.

\item {\bf Theory assumptions and proofs}
    \item[] Question: For each theoretical result, does the paper provide the full set of assumptions and a complete (and correct) proof?
    \item[] Answer: \answerNA{} 
    \item[] Justification: The proposed method does not involve theory assumptions and proof.

    \item {\bf Experimental result reproducibility}
    \item[] Question: Does the paper fully disclose all the information needed to reproduce the main experimental results of the paper to the extent that it affects the main claims and/or conclusions of the paper (regardless of whether the code and data are provided or not)?
    \item[] Answer: \answerYes{} 
    \item[] Justification: We have disclosed all the information, including details of modules, parameters, implementation details. Our code in attached in supplementary materials.

\item {\bf Open access to data and code}
    \item[] Question: Does the paper provide open access to the data and code, with sufficient instructions to faithfully reproduce the main experimental results, as described in supplemental material?
    \item[] Answer: \answerYes{} 
    \item[] Justification: Our code in attached in supplementary materials with detailed instructions and documentations.

\item {\bf Experimental setting/details}
    \item[] Question: Does the paper specify all the training and test details (e.g., data splits, hyperparameters, how they were chosen, type of optimizer, etc.) necessary to understand the results?
    \item[] Answer: \answerYes{} 
    \item[] Justification: We have specified all the training and test details.

\item {\bf Experiment statistical significance}
    \item[] Question: Does the paper report error bars suitably and correctly defined or other appropriate information about the statistical significance of the experiments?
    \item[] Answer: \answerNo{} 
    \item[] Justification: To maintain a similar setting for comparison with previous studies.

\item {\bf Experiments compute resources}
    \item[] Question: For each experiment, does the paper provide sufficient information on the computer resources (type of compute workers, memory, time of execution) needed to reproduce the experiments?
    \item[] Answer: \answerYes{} 
    \item[] Justification: Experiments compute resource information is provided in Implementation section.
    
\item {\bf Code of ethics}
    \item[] Question: Does the research conducted in the paper conform, in every respect, with the NeurIPS Code of Ethics \url{https://neurips.cc/public/EthicsGuidelines}?
    \item[] Answer: \answerYes{} 
    \item[] Justification: The paper adheres to the NeurIPS Code of Ethics.

\item {\bf Broader impacts}
    \item[] Question: Does the paper discuss both potential positive societal impacts and negative societal impacts of the work performed?
    \item[] Answer: \answerYes{} 
    \item[] Justification: Included in the Limitations and Future Work section.
    
\item {\bf Safeguards}
    \item[] Question: Does the paper describe safeguards that have been put in place for responsible release of data or models that have a high risk for misuse (e.g., pretrained language models, image generators, or scraped datasets)?
    \item[] Answer: \answerNA{} 
    \item[] Justification:  There are no released models and scraped datasets.

\item {\bf Licenses for existing assets}
    \item[] Question: Are the creators or original owners of assets (e.g., code, data, models), used in the paper, properly credited and are the license and terms of use explicitly mentioned and properly respected?
    \item[] Answer: \answerYes{} 
    \item[] Justification: We have properly mentioned the used models and cited them without violating their license.

\item {\bf New assets}
    \item[] Question: Are new assets introduced in the paper well documented and is the documentation provided alongside the assets?
    \item[] Answer: \answerYes{} 
    \item[] Justification: The code is with the MIT license.

\item {\bf Crowdsourcing and research with human subjects}
    \item[] Question: For crowdsourcing experiments and research with human subjects, does the paper include the full text of instructions given to participants and screenshots, if applicable, as well as details about compensation (if any)? 
    \item[] Answer: \answerNA{} 
    \item[] Justification: The paper does not involve crowdsourcing nor research with human subjects.

\item {\bf Institutional review board (IRB) approvals or equivalent for research with human subjects}
    \item[] Question: Does the paper describe potential risks incurred by study participants, whether such risks were disclosed to the subjects, and whether Institutional Review Board (IRB) approvals (or an equivalent approval/review based on the requirements of your country or institution) were obtained?
    \item[] Answer: \answerNA{} 
    \item[] Justification: The paper does not involve crowdsourcing nor research with human subjects.

\item {\bf Declaration of LLM usage}
    \item[] Question: Does the paper describe the usage of LLMs if it is an important, original, or non-standard component of the core methods in this research? Note that if the LLM is used only for writing, editing, or formatting purposes and does not impact the core methodology, scientific rigorousness, or originality of the research, declaration is not required.
    \item[] Answer: \answerNA{} 
    \item[] Justification: The core method development in this research does not involve LLMs as any important, original, or non-standard components.

\end{enumerate}